%% file: main.tex
\documentclass{vldb}
\usepackage{graphicx}
\usepackage{balance}  

\newcommand{\ix} {\hspace*{2em}}

\begin{document}




\title{Enabling Cognitive Intelligence Queries in Relational Databases using Low-dimensional Word Embeddings}



%
%
%
%

\numberofauthors{2} 

\author{
%
%
\alignauthor
Rajesh Bordawekar\\
       \affaddr{IBM T. J. Watson Research Center}\\
       \affaddr{Yorktown Heights, NY 10598}\\
       \email{bordaw@us.ibm.com}
\alignauthor
Oded Shmueli\titlenote{Work done while the author was visiting IBM Research.}\\
       \affaddr{Computer Science Department}\\
       \affaddr{Technion, Haifa 32000, Israel}\\
       \email{oshmu@cs.technion.ac.il}
}

\maketitle

\input{abstract}
\input{intro}

\input{related}
\input{vectors}

\input{example}
\input{case-studies}

\input{concl}

\bibliographystyle{abbrv}
\bibliography{BIB/related}  
\input{appendix}

\end{document}

%% file: abstract.tex
\begin{abstract}
We apply distributed language embedding methods from Natural Language Processing to assign a vector to each database entity associated token (for example, a token may be a word occurring in a table row, or the name of a column).
These vectors, of typical dimension 200,  capture the meaning of tokens based on the contexts in which the tokens appear together.
 To form vectors, we apply a learning method to a token sequence derived from the database. We  describe various techniques for extracting token sequences from a database.
The techniques differ in complexity, in the token sequences they output and in the database information used (e.g., foreign keys).
The vectors can be used to algebraically quantify \emph{semantic relationships} between the tokens such as similarities and analogies.

Vectors enable a \emph{dual view} of the data: relational and (meaningful rather than purely syntactical) text.
We introduce and explore a new class of queries
called \emph{cognitive intelligence (CI)} queries that extract information from the database based, in part, on the relationships encoded by vectors. We have implemented a prototype system on top of Spark \cite{Spark} to exhibit the power of CI queries. Here, CI queries are realized via SQL UDFs.  This power goes far beyond text extensions to relational systems due to the information encoded in vectors.

We also consider various extensions to the basic scheme, including
using a collection of views derived from the database to focus on a domain of interest, utilizing vectors and/or text from external sources, maintaining vectors as the database evolves and exploring a database without utilizing its schema. For the latter, we consider minimal extensions to SQL to vastly improve query expressiveness.

\end{abstract}

%% file: intro.tex
\section{Introduction}

Traditionally, relational databases have been used to analyze
enterprise datasets that comprise mostly of well-qualified typed entities
(e.g., character(n), decimal, float, or timestamp). However, over the
years, relational databases have been increasingly used to store and
process free-formed unstructured text data, e.g., customer
reviews, comments to posts, call center interactions,
medical transcriptions, genomic datasets, system logs, etc.
Databases with such unstructured text entities have a significant amount of latent
semantic information; e.g., a word  has a meaning (e.g.,
\texttt{Deep}), a group of words has a meaning (e.g.,
\texttt{Deep Learning}), and finally, even a group of words in a table row
can be viewed to have a meaning (e.g., a person with \texttt{ID} \texttt{110} with
\texttt{Title} of \texttt{Professor} has a job description, \texttt{Deep
  Learning Research}). At present, there is very limited support in
the SQL infrastructure to develop semantic queries that can exploit
semantic relationships between database tables' entities. Although, many databases
support semantic queries either via text extenders that use dictioneries
to identify word synonyms, e.g., DB2 Text
Extender~\cite{cutlip:db2} or using RDF-based
ontologies~\cite{lim:edbt13}, these approaches lack
capabilities for extracting and using latent semantic information from \emph{all}
the database entities.

In this paper, we introduce and explore a new class of queries, called \emph{Cognitive
  Intelligence (CI)} queries, that extract information from a
database based, in part, on the relationships among database entities encoded as vectors produced by a machine
learning method. The vectors are designed such that vectors
corresponding to closely related entities are also close in the
\emph{semantic (vector)} space using a cost metric. We use a distributed
language embedding method from the Natural Language Processing (NLP)
domain to assign a vector to each
database-associated token (e.g., a token may be a word occurring in a
table row, or the name of a column). Vectors may be produced by either learning
on the database itself or using external text, or vector, sources.
These vectors usually have low dimension (200-300) and
capture the meaning of a token based on the contributions of other
tokens in the contexts in which the token appears.

Over the last few decades, a number of methods have been introduced
for computing vector representations of words in a natural language
\cite{bengio:jmlr03}. The methods range from \emph{brute force} learning by various types of
neural networks~\cite{bengio:jmlr03}, to log-linear
classifiers~\cite{mikolov:corr-abs-1301-3781} and to various matrix
factorization techniques~\cite{levy:nips14}.  Lately,
word2vec~\cite{w2v,mikolov:nips13,mikolov:corr-abs-1309-4168,mikolov:corr-abs-1301-3781}
has gained prominence as the vectors it
produces appear to capture syntactic as well semantic properties of
words. There are alternative mechanisms for producing vectors of similar
quality, for example GloVe~\cite{pennington:glove14} (we use word2vec
as the method for constructing vectors from database entities, although,
we could use GloVe as well).  These vectors can  then be used to compute the semantic and/or
grammatical  closeness (e.g., singular/plural) of words as well as test for analogies such as
\emph{a king to a man is like a queen to what?} (answer: \emph{woman}).



In the database context, a natural way of generating vectors is
to apply a word embedding method (in our case, word2vec) to a
token sequence generated from the database. Our scheme has three main phases:
(1) Generating token sequences from the database tables
(\emph{textification or tokenization}), (2) Applying a vector
construction method to the concatenation of these token sequences, and
(3) Executing CI queries using the vectors produced by this method
(once created, these vectors can be reused for answering any future queries). We usually use the
terms \emph{training} or \emph{learning} to describe the operation of
vector construction.

\begin{figure}[htbp]
        \begin{center}
                \includegraphics[height=3.5in]{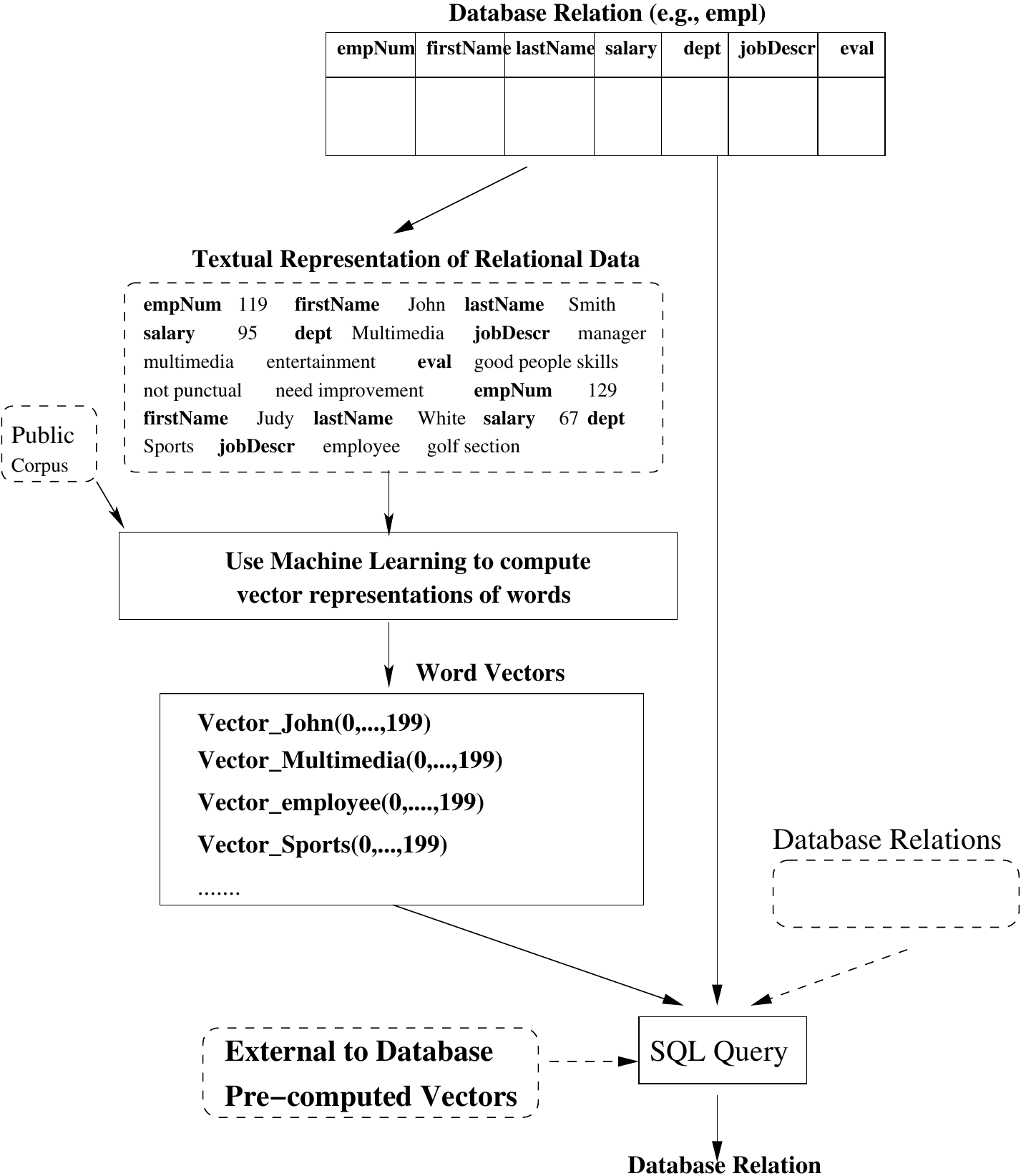}
                \caption{Execution Flow of a Cognitive Intelligence Query}
        \end{center}
\label{fig:flow}
\end{figure}

The execution flow in our scheme is depicted in
Figure~\ref{fig:flow}. At the top, we see a relation, \texttt{empl},
containing rows describing employees. Next, a textual representation
of the data in the \texttt{empl} relation is extracted. Note that, which
rows are textified can be controlled using standard relational
operations (e.g., by defining a view). Next, we use a machine learning  method (e.g., word2vec) to learn vectors for the words (tokens) in the extracted text.
This phase can also use an
external source, e.g., Wikipedia, as the source for text for method
training. The result is a set of low-dimensional (say, of dimension 200) vectors, each
representing one word (token). We use \emph{word} as a synonym to token
although some tokens may not be valid words in any natural language.
In the last phase, these vectors may be used in querying, for example,
via user defined functions (UDFs) in SQL. These UDFs compute
distances between vectors in a semantic (vector) space using a distance metric (e.g., cosine
distance, Jaccard distance) to determine \emph{contextual semantic
  similarities}  between corresponding database
entities. The similarity results are then used to guide the
relational query execution, thus enabling the relational engine to
exploit latent semantic information for answering relational queries.

In this paper, we  discuss various techniques for extracting token sequences from a database.
The techniques differ in complexity, in the tokens they output and in the database information used (e.g., foreign keys).

Vectors can be used to quantify both \emph{syntactic} and
\emph{semantic relationships} between the tokens such as similarities
and analogies. By computing vector averages, database entities such as
a column value in a row,  a whole row,  or even in some cases a whole column or relation,
may be usefully encoded as vectors. This basic scheme can be extended in
multiple ways, e.g., we can first form a collection of views derived from the database to focus
on a domain of interest, and apply vector construction methods to the token sequence
derived from these views. For example, suppose we have a database
containing corporate data. We can restrict it (via a collection of
views) to only information that is related to mid-size companies. Or
in a medical database, restrict to only diabetes related
information. This mechanism restricts the vocabulary (set of tokens)
and focuses training on the domain of interest.

Vectors enable a \emph{dual view} of the data: relational and
(meaningful) text. The text view of a database can  be queried
directly, but more importantly, it can be used for enabling novel CI queries. Such queries can also
\emph{navigate} database tables without precise schema knowledge.  For example, one can ask: list database rows (of any relation) that
have a field whose content designates an address that is physically
nearby to the address of employee 55; the list is ordered by nearness to the
employee 55 address. With this query, we are totally oblivious to
the schema of the database. In fact, we can ask which relations
contain such rows, and rank them according to the number of such rows. Specific examples of this feature are shown in Section \ref{sec:example}.

We  have implemented a prototype
system (on top of Spark SQL \cite{Spark}) to exhibit the power of CI
queries. CI queries take relations as input and return relations as
output.  CI queries are realized via SQL and user defined functions,
UDFs.  CI queries power goes far beyond text extensions to relational systems
due to the information encoded in vectors. For example, we can encode
a field of a row as a vector and perform approximate semantic equality joins on
such fields.  CI queries augment the capabilities of the
traditional relational OLAP queries and can be used in conjuction with
the existing SQL operators (e.g., OLAP~\cite{gray:olap}).

We believe that this is the first work to explore applications of NLP-based machine learning
techniques for enhancing and answering relational queries.
The paper makes the following key contributions:

\begin{enumerate}

\item
Database tokenization and the \textit{association of a vector with each
database entity represented by a token}. This enables a dual
database view: relational and meaningful text.


\item
Introducing a new class of queries, Cognitive Intelligence (CI)
queries, that take advantage of this dual view. CI queries can be
realized within standard SQL using UDFs or with minimal extensions to
SQL. The vectors capture latent semantic relationships among database
text entities.

\item
CI queries enrich SQL's expressiveness and enable sophisticated
approximate similarity, semantic contextual matching and analogy queries.

\item
Using externally computed vectors, the CI queries can evaluate
semantic relationships of entities that \emph{are not even present} in a database
with the database entities.

\item
CI queries can be used to navigate through the database using standard
SQL or a minimal extension thereof, with very little knowledge of the
database schema.

\item
We exhibit the feasibility of the approach by implementing a prototype
system on top of Spark.

\end{enumerate}

 The paper is organized as follows:
First, Section \ref{sec:related} discusses related work.
 Section \ref{sec:vectors} presents various methods for  tokenization -- producing the
 sequence of tokens for training, using external sources for text and
 vectors and operational issues.  Section \ref{sec:example} presents usage examples.
 In Section \ref{subsec:similarity},  we present similarity queries.
 In Section \ref{sec:navigation}, we explore query language extensions
 to support database navigation with minimal schema knowledge.
 In Section \ref{subsec:analogy}, we briefly introduce the power of analogy
queries.
Section \ref{sec:case-studies} presents initial experiences of executing CI queries using
 a publically available dataset (DBLP), and outlines case-studies from other relevant domains.  
 It also describes a prototype implementation on top
 of Spark SQL.  An appendix reviews word2vec, the system we used to generate
 vectors, as well as mentions other similar systems.
 We conclude in Section \ref{sec:concl}.

%% file: related.tex
\section{Related Work}
\label{sec:related}
\textbf{Language Embedding:}
Over the last 25 years or so, a number of methods have been introduced for obtaining a vector representation of words in a language \cite{bengio:jmlr03}, called \textit{language embedding}.
The methods range from "brute force" learning by various types of
neural networks (NNs) ~\cite{bengio:jmlr03}, to
log-linear classifiers~\cite{mikolov:corr-abs-1301-3781} and to
various matrix formulations, such as matrix  factorization techniques
\cite{levy:nips14}.

Lately, word2vec~\cite{w2v, DBLP:Rong14-v2} has
gained prominence as the vectors it produces appear to capture
syntactic as well semantic properties of words. word2vec is a popular
machine learning method primarily used in the field of Natural
Language Processing (NLP)~\cite{mikolov:nips13, mikolov:corr-abs-1309-4168, levy:conll14}. Scanning a corpus (text), it
generates a vector representation for each word in the text. A
vector is usually of a low dimension (about 200-300) and represents
the word. The vectors can be used to compute the semantic and/or
grammatical closeness of words as well as test for analogies.
%
%
The exact mechanism employed by word2vec and suggestions for alternatives are
the subject of much research~\cite{mnih-nips13, goldberg:corr-abs-1402-3722, levy:conll14, pennington:glove14, arora:corr-abs-1502-03520}.
We note therefore that although  word2vec has gained much prominence
it is one of many possible methods
for generating word representing vectors.
Additional details concerning word2vec are provided in the appendix.

Vectors may be associated with larger bodies of text such as paragraphs and even documents. Applications to the paragraph and document embedding appear in \cite{le:corr14, dai:corr15}. Recent work has also been exploring applying word embeddings to capture image semantics~\cite{socher:zero-shot}.


\textbf{Relational Databases:}
In the context of SQL, text capabilities, e.g., the use of synonyms,
have been in practice for while \cite{cutlip:db2}.
In the literature, techniques for detecting similarity between records and fields have also been explored, some of the relevant references follow.
Semantic similarity between database records is taken into account
in~\cite{kashyap:vldbj96}.
Phrase-based ranking by applying an IR approach to relational
data appears in~\cite{liu:sigmod06}.
Indexing and searching relational data by modeling tuples as virtual
documents appear in~\cite{luo:sigmod07}.
Effective keyword-based selection of relational
databases is explored in~\cite{yu:sigmod07}.
A system for detecting XML similarity, in content and structure, using
a relational database is described in~\cite{viyanon:cikm09}.
Related work on similarity Join appears in~\cite{chaudhuri:ssjoin}.
Semantic Queries are described in~\cite{pan:dldb, lim:edbt13}.
Knowledge Databases \cite{vilnis:corr-1412-6623,
  verga:corr-1511-06396} are relevant to extracting semantics. Most
recently, Shin et. al have described DeepDive~\cite{Shin:vldb} that
uses machine learning techniques, e.g., Markov Logic based rules, to
convert input unstructured documents into a structured knowledge base.

What distinguishes this work from the relevant prior work is that we
enable queries on relational data that exploit latent semantic
information in the relational database. We use NLP techniques for associating
 each database text entity with a vector that captures its
syntactical and semantic relationship to other database text
entities. Further, these vectors are primarily based on the database \textit{itself}
(with external text or vectors as an option). This means that we
assume no reliance on dictionaries, thesauri, word nets and the
like. Once these vectors are produced (using machine learning methods)
they may be used in vastly enriching the querying expressiveness of virtually
any query language.

%% file: vectors.tex
\section{Tokenization}
\label{sec:vectors}
\subsection{Basic Tokenization}
Distributed language embedding refers to assigning a vector to each
word, term, concept, or more generally token, appearing in a sequence where the vectors
indicate various aspects of the associated words, enabling computing semantic
\emph{closeness}. Distributed language
embedding  can expose \emph{hidden}  entity relationships in structured
databases. The term \emph{text entity} is used  to refer to some
discernible item appearing in a database (or some external source),
such as a character, a delimited character sequence, a Boolean value,
a number, or a meaningful short sequence such as \emph{Theory of
Relativity}. The text entities are determined by a tokenization
process that takes the content of a row field and identifies a
sequence of one or more text entities within it.
The tokenization process then  uniquely associates a token with each
distinct  entity.
Technically, a token is a string with no intervening white space or
punctuation  and is also referred to as a word.
Here are some possible example tokens: \texttt{TRUE}, \texttt{USA}, \texttt{physical},
\texttt{Theory\_of\_Relativity}.

A \emph{word vector} is a vector representation of a token (word) in
a language whose vocabulary is the tokens.
The methods for obtaining  vector representations
range from 'brute force' learning by various types of neural networks,
to log-linear classifiers and to various matrix formulations, such as
matrix factorization techniques.
One example method is a tool called
‘word2vec’ which produces vectors that capture syntactic as well as
semantic properties of words. word2vec scans a corpus (text) to
generate vector representations for each word in the text. A word
vector is usually of a low dimension (200-300) and represents
the word (token). All vectors trained on a sequence of tokens have the \emph{same dimension}.
Closeness of vectors is determined by using the cosine  or Jacard distance measures
between vectors (other measures may apply for specific applications).
The vectors can be used to compute the semantic and/or
grammatical closeness of words as well as test for analogies (such as \emph{a
king to a man is like a queen to what?}).
Sections \ref{sec:related} and the appendix briefly survey the
language embedding field (i.e., vector representation of words).

\textbf{Tokenizing a database:} A database is a collection of tables.
Each table is a collection of rows having the same number of columns.
Each column is associated with a data type of the usual character, string, Boolean and numeric types, or BLOB (which we do not address in this paper).
The value of a row in a particular column conforms with the data type of the column.
Vector learning devices operate on (text) documents (for us the token
sequence is the document). So, the database needs to be \textit{tokenized} so
as to form a token sequence (the document).
The basic tokenization process operates as follows.
A character value such as '\texttt{y}' is represented by the string ``\texttt{y}''.
A Boolean is represented as ``\texttt{TRUE}" or ``\texttt{FALSE}".
Numbers are represented in their textual representation, for example,
the integer \texttt{12} is represented as the token ``\texttt{12}", and the
real number \texttt{123.001} by the token  ``\texttt{123.001}". A Null is represented as ``\texttt{Null}".

A string value is tokenized as a string containing a sequence of words separated by blanks;
these words are extracted by the tokenization process from the input string value.
There are a few \emph{tokenization options} as to how this extraction is performed.
For example, viewing a sequence of words such as ``\texttt{Deep
  Learning}" as a single token \newline ``\texttt{Deep\_Learning}".
A row is tokenized as a token sequence made by concatenating the sequences of tokens
tokenized from its fields (columns) in order. A table is tokenized as a token
sequence made by concatenating the sequences of tokens  tokenized from its rows, and the whole
database is tokenized by concatenating the token sequences of its
tables (again, there are a few options here).

\begin{figure*}
        \begin{center}
                \includegraphics[height=2in]{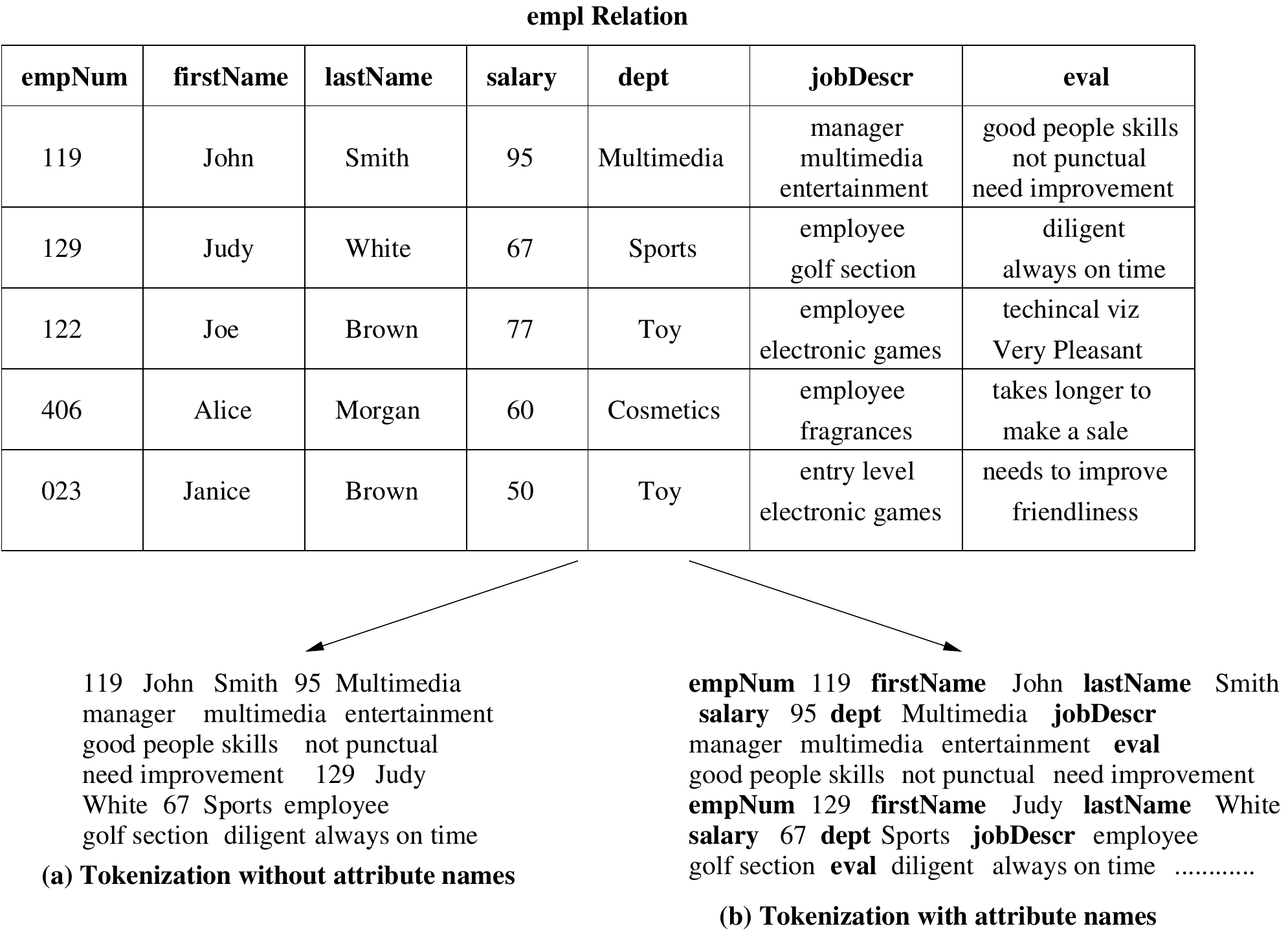}
                \caption{Dual View of a Relational Table}
        \end{center}
\label{fig:dual-view}
\end{figure*}

The  token sequences in Figure 2 (a), and (b) exhibit two fundamental ways
 in which  token sequences
may be derived from a database.  For example, the tokens
corresponding to John Smith's jobDesc (i.e., ``\texttt{manager
multimedia entertainment}" in the sequence of tokens) may instead
be derived as ``\texttt{jobDesc manager jobDesc multimedia jobDesc
entertainment}", in which case the resulting trained vectors would
likely show a stronger relationship between the vector of word
``\texttt{jobDesc}" and the word vectors of ``\texttt{manager}", ``\texttt{multimedia}", and
``\texttt{entertainment}".

\subsection{Extensions}

\textbf{Encoding Relation Name:} There are some extensions to these basic tokenization techniques.
In Figure 2 (a) each field tokens are preceded by
the field's name. Similarly, one may precede each row of a relation
with a token uniquely representing the associated relation. Yet a
third possibility is to precede some token sub-sequences of a field
with both relation name and field name.

\textbf{Support for foreign keys:} A foreign key in a relation
A with respect to a relation B is a set of columns in A whose values
uniquely determine a row in B. When a foreign key is present, during
token generation for relation A, we can \emph{follow} the foreign key to a
row in relation B. We can then tokenize fields of interest in the row of relation
B and insert the resulting sequences into the sequence generated for
relation A.
Figure 3 presents another example of a
database table, \texttt{address}, and a resulting token
sequence that utilizes a relationship between the \texttt{empl} relation and the \texttt{address}
relation; namely the \texttt{address} table
provides the addresses for the employees of database table \texttt{empl}.
Technically, the resulting token sequence is based on foreign key
\texttt{empNum} in \texttt{emp} which provides a value for key attribute \texttt{id} of the
\texttt{address} relation. The straight forward way to tokenize with foreign
keys is to insert the subsequence generated out of the B row
immediately after the one generated for the A row as depicted in
Figure 3; another possibility is to intermix the
subsequence from the B row within the A row sequence following the
tokenization of the foreign keys values of the A row (again, other
options may apply).

\begin{figure*}
        \begin{center}
                \includegraphics[height=1.5in]{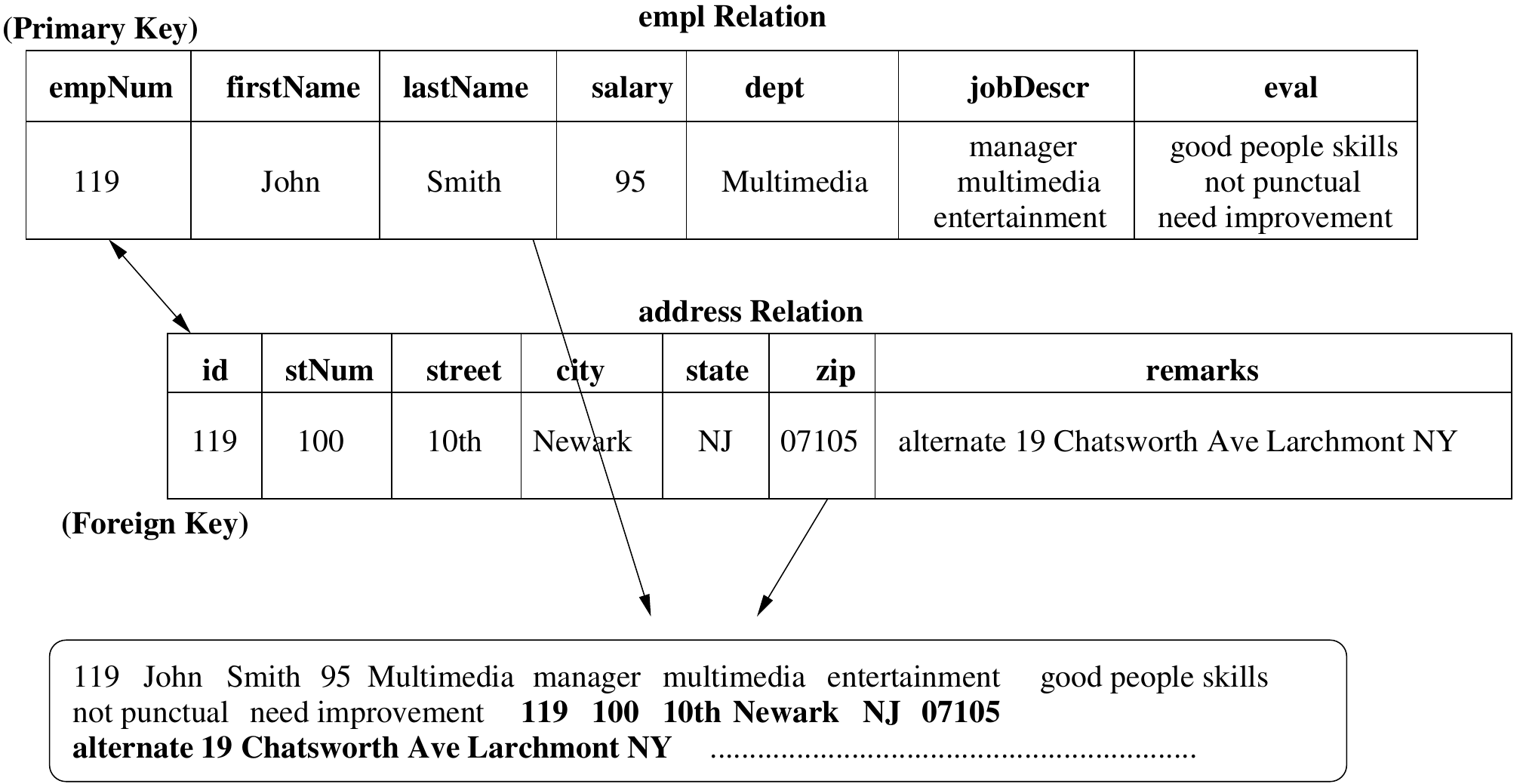}
                \caption{Text view of two tables joined using primary and foreign keys}
        \end{center}
\label{fig:foreign}
\end{figure*}

Using foreign keys in
generating the set of token sequences provides a user with the ability
to make connections between entities of database tables that are not easily
evident in the database. For example, if many news articles
mention Mamaroneck and Larchmont together then the vectors of the
corresponding tokens will be close. If a user constructs a query to
locate employees that live close to each other and Alice Morgan lives at
9999 Main Street, Mamaroneck, and Janice Brown lives at 1000
Hutchinson Ave Larchmont, then \texttt{proximityMAX(9999 Main Street Mamaroneck, 1000 Hutchinson Street Larchmont)} is high because the
cosine\_distance($V\_{Mamaroneck}, V\_{Larchmont}$) is high. \newline Given
two sets of vectors, the UDF \texttt{proximityMAX()} returns the highest cosine distance
between a vector from one set and a vector from the other set.

\textbf{Support for numerical values:} A third extension concerns numerical values.
In this extension, each number is also preceded by a \emph{range
  designator}.   Designators are system parameters. They allow
detecting closeness of numerical values whose textual representation
may not yield close vectors despite their numerical closeness. For
example, for positive numbers we may consider ranges such as ``\texttt{1-4, 5-9,
10-49, 50-99, 100-499, 500-999, 1000-4999}",... Similar ranges may be
constructed for values between 0 and 1 and for negative
numbers. Continuing the example, if a field contains the number \texttt{78.5} the
corresponding output sub-sequence of tokens would be ``\texttt{50-99 78.5}". For
lack of space, we shall not display designators in examples and
figures.

A fourth extension concerns using external token sequences (text) as
well as externally produced vectors. Tokenizing the database concatenated (in various ways)
with an external token sequence has the advantage that learning is
performed on a larger document with possibly a larger vocabulary. A
larger document usually means better learning; a larger vocabulary
\emph{enhances} user querying abilities and enables querying with text
entities that do not even appear in the database.
One can even contemplate using two (or more) sets of vectors where in
each usage (e.g., \texttt{proximityMAX()}), the relevant vector set is
identified. Yet another possibility is to train on the token sequence
of the database with vectors for tokens that also appear in the
external source \emph{fixed} throughout the vector  training process.

The above described tokenization methods may be extended to non-relational
 databases as well. They can be applied to JSON, XML, RDF (and the
query language to be extended may be one of  Xpath, Xquery, SPARQL, HQL,
Spark SQL). Generating token sequences for tree and graph structures is more complex.
 The basic idea is exploring \emph{relevant} token sequences such as those
 implied by paths, by siblings, and combinations thereof. We will not
 elaborate further on this due to space limitations.

\textbf{Mainainance of Vectors:} As the database changes vectors may
need adjustment. One option is periodical retraining that can be
applied to a snapshot of the database. Another option is continuous
adjustments and handling of brand new tokens. We have explored
techniques for continuous adjustments and due to space limitations
defer discussion to a future paper.

%% file: example.tex
\section{Cognitive Intelligence Queries By Examples}
\label{sec:example}

\subsection{Similarity Queries}
\label{subsec:similarity}

In this section we present examples from two domains:  a business
Human Resources (HR) database and a scientific publications database.
In both domains we use SQL queries that invoke user defined functions
(UDFs).  Both UDFs take two arguments and return a real value between -1.0 and
1.0.

\texttt{proximityMax()} is a UDF that (a) generates two sets of tokens S1 and S2, S1
from its first and S2 from its second argument,  (b) calculates the
cosine distance of the vector of each token in S1 with the vector of
each token in S2, and (c) returns the maximum such cosine distance (a
real number between -1.0 and 1.0). Intuitively, \texttt{proximityMax()}  assesses the
closeness of two token sets by the maximum closeness of an
element (token) from one set and an element (token) in the other set
using the bag-of-words model ~\cite{manning:ir}. In case either set is
empty, -1.0 is returned.

\texttt{proximityAvg()} is a UDF that (a) generates two sets of
tokens, one from its first and one from its second argument, (b)
represents a set of tokens via a single vector which is the
average vector of the vectors associated with the tokens in the set, and
(c) returns the cosine distance of the two vectors representing the
two sets.

We note that both \texttt{proximityMax()} and \texttt{proximityAvg()} ignore highly frequent
tokens such as : \texttt{on}, \texttt{up}, \texttt{down}, \texttt{a}, \texttt{the}, \texttt{their}, \texttt{its}, \texttt{if},
\texttt{his}, \texttt{her}, \texttt{and}, \texttt{or}, \texttt{not}, \texttt{of}, \texttt{in}, \texttt{for}, \texttt{using} as they
provide little context information (but may still be useful in the
vector learning process).

 Consider  a publication relation with columns number, author, title,
 year where number is key and the columns have their obvious
 meaning. At this point we shall focus on querying and delay
 discussing the tokenization process  till the next example. So assume
 the publication database was already tokenized and each database text
 entity is associated with a vector (of its token).
Suppose we would like to list pairs of authors, author 1 and author 2,
and publications that deal with similar topics. We quantify \emph{similar}
by using \texttt{proximityMax()} to determine if there is at least one token
derived from a title of an author 1 article that is very close to a
token derived from a title of an author 2 article. Here \emph{very close} is
quantified as cosine distance greater than 0.3. The following query
operates over Spark SQL.

\begin{figure}[htbp]
\hrule
{\small \tt{
\ix \\
 SELECT X.number, X.author, X.title, Y.number, Y.author, \\
 Y.title, proximityMax(X.title, Y.title) AS proximityMax \\
 FROM papers X, papers Y \\
 WHERE int(X.number) < int(Y.number) AND\\
 proximityMax(X.title, Y.title) > 0.3 \\
 LIMIT 10\\
}
\hrule
}
\caption{Example of using the \texttt{proximityMax()} UDF}
\label{fig:example1}
\end{figure}

The example query above returns up to 10 results, where each result
will list \texttt{number1, author1, title1, number2, author2, title2}, \texttt{proximityMax()} value such
that \texttt{title1} and \texttt{title2} are concerned with at least one very similar
topic. The statement \texttt{int(X.number) < int(Y.number)} prevents a row
from being compared to itself and from a pair of qualifying authors to
appear twice in the result due to the same two articles. 

Let us
examine how the statement \texttt{proximityMax(X.title, Y.title) $>$
  0.3} is processed. Suppose \texttt{X.title}= \texttt{Linear \newline
  Approximation of Image
Similarity} and \texttt{Y.title} = \newline \texttt{Examining Algebraic Identification
Methods}. Then, a standard tokenization will identify the tokens
\texttt{Linear}, \newline \texttt{Approximation}, \texttt{of}, \texttt{Image}, \texttt{Similarity} in X.title and
\texttt{Examining}, \texttt{Algebraic}, \texttt{Identification}, \texttt{Methods} in
Y.title. Next, two sets of vectors will be identified V1 = \{
$V_{Linear}$, $V_{Approximation}$,  $V_{Image}$,  $V_{Similarity}$\},
and V2= \{ $V_{Examining}$, $V_{Algebraic}$, \newline
$V_{Identification}$\}. There is no $V_{of}$ as \texttt{of} is excluded due
to its high frequency of occurrence.  Next, the cosine distance is
calculated between each member of V1 and each member of V2. If any of
these calculations is higher than 0.3, tuples X and Y qualify
(provided $X.number < Y.number$).

One may argue that \texttt{proximityMax()} is not appropriate as it
does not take the whole title into account. As an alternative, one can
use another UDF, \texttt{proximityAvg()}:

\begin{figure}[htbp]
\hrule
{\small \tt{
\ix \\
SELECT X.number, X.author, X.title, Y.number, Y.author, \\
Y.title, proximityAvg(X.title, Y.title) AS proximityAvg \\
FROM papers X, papers Y \\
WHERE int(X.number) < int(Y.number) AND\\
proximityAvg(X.title, Y.title) > 0.2 \\ 
LIMIT 15\\
}
\hrule
}
\caption{Example of using the \texttt{proximityAvg()} UDF}
\label{fig:example2}
\end{figure}

The example query above (also in Spark SQL) returns up to 15 results,
here each result lists \texttt{number1, author1, title1, number2, author2, title2,}
\texttt{proximityAvg()} value such that \texttt{title1} and \texttt{title2} generally deal with similar
topics. Note that a lower bound of 0.2 is used instead of 0.3 in the
previous query as the average correlation will tend to be lower in
this scenario. Note also that excluding commonly occurring terms will
enhance accuracy. In general, the exact set of terms to be excluded
could also be a parameter. Also,  splitting a string on spaces (blanks)
may be changed to splitting of other forms, e.g. on tabs, commas.

One can define additional UDFs, for example a variation on
\texttt{proximityMax()}, \texttt{proximityTop2Avg()}, that returns the average of the top 2
cosine distances rather than the maximum,  UDFs that use a different
distance measure than cosine distance, UDFs that return minimum rather
than maximum cosine distances and so forth. 

We next continue with an examples from the HR domain.
Figure 2 displays five records of employee information from table
 emp; emp  has the following seven columns (fields): \texttt{empNum,
firstName, lastName, salary, dept, jobDesc}, \newline and \texttt{eval} (containing free text
employee evaluation); the column names are self-explanatory.
Figure 2 (a) and (b)  illustrate two simple examples
of resulting tokenizations of these database tables.
The idea is to convert each row to text and concatenate these texts. A
row is converted one column at a time.
The exhibited  token sequences (a) and (b) each  present a possible
tokenization of the five records of employee information
stored in the database emp table.
The difference between (a) and (b) in the figure is that column
(field) names also appear as text just ahead of their content in (b). 

In Figure \ref{fig:flow} the resulting sequence of tokens,
(a) or (b), in this example, is presented to a machine learning
device (such as word2vec) to obtain a vector for each token
in the sequence. Optionally, other information may be fed into the
machine learning device, for example a public corpus (text) such as
Wikipedia. The machine learning device  then outputs vectors
representing the entities in the database table. The
database entity \texttt{John} in the example shown in Figure 2 is assigned a
vector of 200 dimensions, denoted by the vector $V\_John(0,…199)$.
However, in other situations a different dimension value may
be used (same dimension for all vectors).

Typically, if a user would like to gather information, say financial,
regarding an employee, then the user would have to be
familiar with the attribute names of the database tables. For example,
if a user wanted to gather information about employee’s salaries, then
the user need construct a SQL query which specifically states the
attribute name \texttt{salary}. However, by incorporating  vectors  a
user gains the ability to construct queries with terms that \emph{are not even in
the database} when the training is based on a rich external source (optionally with the database), such
as Wikipedia. This will create the \emph{connection} between terms mentioned in the database and ones that are not.

For example, suppose we produce a token
sequence as in Figure 2 (b), i.e., including column
names. The sequence of tokens  would be \texttt{empNum, 119,
firstName, John, lastName, Smith, salary, 95, dept, Multimedia,  jobDesc,
manager, multimedia, entertainment, eval, good, people, skills, punctual,
need, improvement,} \ldots . Further, suppose that other relations include
columns such as \texttt{bonus}, \texttt{fine}, \texttt{sales} and the like that mention \texttt{Smith} and that are used in training. Then, the user
may construct a query regarding Smith's financial records, in a history relation, by simply
stating the query in Figure \ref{fig:example3}.  

\begin{figure}
\hrule
{\small \tt{
\ix \\
SELECT * \\
FROM history h \\
WHERE contains(h, "Smith") \\
AND proximityMax(h, "money") > 0.5 \\
}
\hrule
}
\caption{Example of semantic similarity}
\label{fig:example3}
\end{figure}

Although the token \texttt{money} may not even be mentioned at all in the history relation, any
tuple in the history relation that mentions such text entities as \texttt{Smith, salary, sale,
bonus, fine, IRA} will likely be retrieved because of the associated
 closeness of the vectors of these text entities and the vector for \texttt{money}
 as will likely be established by the training process. \emph{This
type of functionality is not currently available to relational database users.}

A typical use of Similarly queries is shown in the following example
in which one  inquires for  job evaluations of employees that are
similar to that of \texttt{Smith}:

\begin{figure}[htbp]
\hrule
{\small \tt{
\ix \\
SELECT * \\
FROM emp e, f \\
WHERE e.lastName = "Smith" \\ 
AND proximityAvg(e.eval, f.eval) > 0.5\\
}
\hrule
}
\caption{Example of semantic similarity using external entities}
\label{fig:example4}
\end{figure}

To understand the exploitation of foreign keys, consider the following
example. Suppose we have relation \texttt{sales} that describes sales, and
has, among other columns, the columns \texttt{authorizedBy} and \texttt{followupBy}. In
both these columns there are employee numbers of the relevant
personnel, moreover these employee numbers are foreign keys into the
\texttt{empl} relation whose primary key is \texttt{empNum}. A tuple in
\texttt{sales} that does not even mention \texttt{Smith} may 
well be retrieved by a query that inquires about \texttt{Smith} if
this tuple is connected to \texttt{Smith} via a foreign key (or
keys). This is due to the fact that when the token sequence is produced using foreign keys
(as discussed previously in further detail with respect to
Figure 3), the token \texttt{Smith} will be associated 
with tokens in tuples that are connected
to the tuple containing \texttt{Smith} via one or more \emph{hops} through
relationships implied by following foreign keys.
So, a user may query the \texttt{sales} relation for sales made by, or related
to, \texttt{Smith} even though \texttt{Smith} may not explicitly be mentioned in the
sales relation (Figure~\ref{fig:example5}).

\begin{figure}[htbp]
\hrule
{\small \tt{
\ix \\
SELECT e.* \\
FROM sales e \\
WHERE proximityMax(e, "Smith") > 0.1 \\
AND proximityMax(e, "money") > 0.5\\
}
\hrule
}
\caption{Example of a semantic similarity query over relations joined
  by primary/foreign keys}
\label{fig:example5}
\end{figure}

The UDF \texttt{proximityAvg()} can be biased due to long token
sequences which lead to mixing many vectors and the
UDF \texttt{proximityMax()} can be biased due to having one strong
interaction and other very weak ones.
Another UDF, \newline \texttt{subsetProximityAvg(size, sequence1, sequence2)} can circumvent these
problems. This UDF takes an additional integer parameter, size. Intuitively, it considers all sets
of tokens out of sequence1 of cardinality size and compute an average
vector for each such subset. Similarly, it does the same for subsets
of tokens of cardinality size taken out of sequence2. It then computes
the cosine distances between each average vector associated with
sequence1 and each average vector associated with sequence2, and
returns the maximum value. So, intuitively, it returns 'maximum of
cosine distances between average vectors of subsets'. This places the
\texttt{subsetProximityAvg()} in between \texttt{proximityMax()} and
\texttt{proximityAvg()}. \texttt{subsetProximityAvg()} is a
generalization of the \texttt{proximityTop2Avg()}.

\subsection{Schema-less Navigation}
\label{sec:navigation}

We now consider \emph{extensions} to SQL that take advantage of the existence of vectors associated with database entities.
The first extension is an ability to focus on individual entities within a field of a row.
We add entities as \emph{first class citizens} by adding a phrase \texttt{Entity e} in the \texttt{FROM} part of a SQL query.
We then extend SQL with the constructs (a) \texttt{contains(column, entity)},
(b) \texttt{contains(row, entity)}, (c) \texttt{contains(database, entity)},(d)
\texttt{cosineDistance( \newline entity1, entity2)} $RELOP$ $c$, where \texttt{entity, entity1},and \texttt{entity2}
are token denoting variables or token constants and $c$ denotes a
constant $ -1.0 \leq  c  \leq  1.0$, $RELOP$ is one of \{$ \leq , \geq , > , < , =      $\}.

The declaration \texttt{Token entity e } declares \texttt{e} to be a variable that can be bound to tokens.
The statement \texttt{contains(\newline row, entity) } states that $entity$ must
be bound to a token generated by tokenizing $row$.  If $entity$ is
specified, e.g., \texttt{contains(r, "Physics")} then the result is TRUE
for row $r$ whose tokenization includes the token \texttt{Physics} and FALSE
otherwise. The argument $column$ is specified by a relation name or
alias followed by the column name as in
\texttt{contains(e.jobTitle, "manager")} where \texttt{e} is an alias for the
\texttt{emp} relation. To indicate a whole row we use the star
notation as in \texttt{contains(e.*, "manager")}. 
In fact, the function \texttt{contains()} returns the number of
occurrences where a number greater than zero is interpreted as TRUE. 
One can use the returned value in expressions such as
\texttt{contains(e.*, "expert") $>$ 1}.

The ability to specify entities provides for a finer query
specification as compared to the UDFs we saw thus far. Relations may
be navigated using entities in a similar way as database keys. For
example, the following query (Figure~\ref{fig:sql-entity}) checks for the existence of two strongly
related (cosine distance $>$ 0.5)  entities in the address field of an
employee and a row (tuple) in the \texttt{DEPT} table (relation).

\begin{figure}[htbp]
\hrule
{\small \tt{
\ix \\
SELECT EMP.Name, EMP.Salary, DEPT.Name \\
FROM EMP, DEPT, Token e1, e2\\
WHERE contains(EMP.Address, e1) AND\\
contains(DEPT.*, e2) \\
AND cosineDistance(e1, e2) > 0.5\\
}
\hrule
}
\caption{Example of an SQL query with entities}
\label{fig:sql-entity}
\end{figure}

To enhance querying ability, we can proceed one step further and
introduce \emph{relation variables} whose names are not specified at
query writing time; they are identified at run-time via entities.
Introducing relation variables is done as shown in the following
example.

\begin{figure}[htbp]
\hrule
{\small \tt{
\ix \\
SELECT EMP.Name, EMP.Salary, S.X\\
FROM EMP, DEPT; Token e1, e2; Relation S; column X\\
WHERE contains(EMP.Address, e1) AND\\
contains(S.X, e2) AND\\
cosineDistance(e1, e2) > 0.5 \\
AND contains(S.X, e2) > 1  \\
}
\hrule
}
\caption{Example of an SQL query with relation variables}
\label{fig:sql-relation}
\end{figure}

In this example query (Figure~\ref{fig:sql-relation}), field (attribute) values are retrieved from \texttt{EMP}
and the (unknown) relation \texttt{S} and column(s) \texttt{X} such that there are
entities \texttt{e1} and \texttt{e2} that strongly \emph{link} them. The \emph{related} relation
and column names and the relevant values may also be
retrieved. Intuitively, we retrieve names and salaries from \texttt{EMP} and
tuple columns \texttt{X} out of some other relation \texttt{S} such that
(the tokenization of)  \texttt{EMP.Address} contains a token  \texttt{e1},
(the tokenization of) \texttt{S}'s column \texttt{X} contains a token \texttt{e2}, and \texttt{e1}
and \texttt{e2} are \emph{close} (cosine distance $>$ 0.5) and has more than 1
occurrence of  token \texttt{e2}. In general, there may be more than one
column in a qualifying relation that qualifies and there may be more
than one qualifying relation, and as such, the construct \texttt{S.X} should
indicate the relation and column names. For example, a result tuple
may look like:
 \texttt{(‘John Smith’, 112000, Dept.Mgr:‘Judy Smith’)}.

The closeness  between entities may also be expressed qualitatively on
a scale, for example \emph{very\_strong, strong, moderate, weak,
  very\_weak}, e.g. \emph{strong(e1,e2)}. which enables defining the
numeric values separately, e.g. $very\_strong = 0.95$. For example,
understanding the relationship between the two text entities \texttt{John} and
\texttt{New York} includes deriving the tokens for the  entities, in this
case simply \texttt{John} and \texttt{New\_York}, locating their associated vectors,
say $V_{John}$ and $V_{New\_York}$, and performing an algebraic operation
on the two vectors, for example, calculating the cosine distance. If
the cosine distance between these vectors is high (i.e., closer to 1)
then the two text entities are closely related; if the cosines distance is
low (i.e. closer to -1) the two entities are not closely related.

We note that the notation \texttt{S.X} is basically syntactic sugar. A software
translation tool can substitute for \texttt{S.X} actual table name and
column. Then, perform the query for each such substitution and return
the union of the results.

\subsection{Analogy Queries}
\label{subsec:analogy}

A unique feature of vectors is the fact that we can use them to deduce
analogies; e.g., \emph{a king to man is like a queen to woman}. The
following query (Figure~\ref{fig:analogy}) identifies pairs of
products that relate to each other like tokens \texttt{peanut\_butter} and \texttt{jelly} (both are
product types). The key is that if product $p1$ relates to product $p2$ like
peanut butter to jelly, their associated vectors $V\_{p1}$, $V\_{p2}$,
\newline $V\_{peanut\_butter}$ and $V\_{jelly}$ are likely to satisfy that the vector
differences ($V\_{peanut\_butter}$ - $V\_{jelly}$) and ($V\_{p1}$ - $V\_{p2}$) are
cosine-similar. Potential answers to this query can include (\texttt{chips},
\texttt{salsa}) or (\texttt{pancake}, \texttt{maple\_syrup}). Here, we use a UDF \texttt{vec()} that takes a token and returns
its associated vector.

\begin{figure}[htbp]
\hrule
{\small \tt{
\ix \\
SELECT P1.Name, P2.name \\
FROM Products P1, Products P2, Token e1, e2\\
WHERE P1.name < P2.name AND\\
cosineDistance(vec(P1.type), vec("peanut\_butter") - vec("jelly") + vec(P2.type)) = \\
  (SELECT  MAX (cosineDistance(vec(P3.type), vec("peanut\_butter") - vec("jelly") + vec(P2.type)))\\
   FROM  Products P3) \\
}
\hrule
}
\caption{Example of a CI analogy query}
\label{fig:analogy}
\end{figure}

In the analogy query we assume that column $type$ of the relation $Product$ contains a single
word, in case it may contain more than one word we can obtain the
tokens within them using an UDF \texttt{contains()}. We note that, assuming all vectors are of length 1, instead of the traditional
{\small\tt{ cosineDistance(vec(PX.type), vec(P1.type) - \newline vec("peanut\_butter") + vec("jelly")))}}, 
we can also use the 3COSMUL formulation of \cite{levy:conll14}, namely:\\
{\small\tt{(cosineDistance(vec(PX.type),vec(P1.type) * }}\\ 
{\small\tt{cosineDistance(vec(PX.type),vec("jelly"))) / }}\\
{\small\tt{(cosineDistance(vec(PX.type),vec("peanut\_butter")) + 0.001)}}. 

%% file: case-studies.tex
\section{CI queries in practice}
\label{sec:case-studies}

In this section, we discuss how a Cognitive Intelligence (CI) database is
used in practice, and then demonstrate its capabilities by describing
execution of CI queries in a Spark-based prototype.

\begin{figure}[htbp]
        \begin{center}
                \includegraphics[height=1.5in]{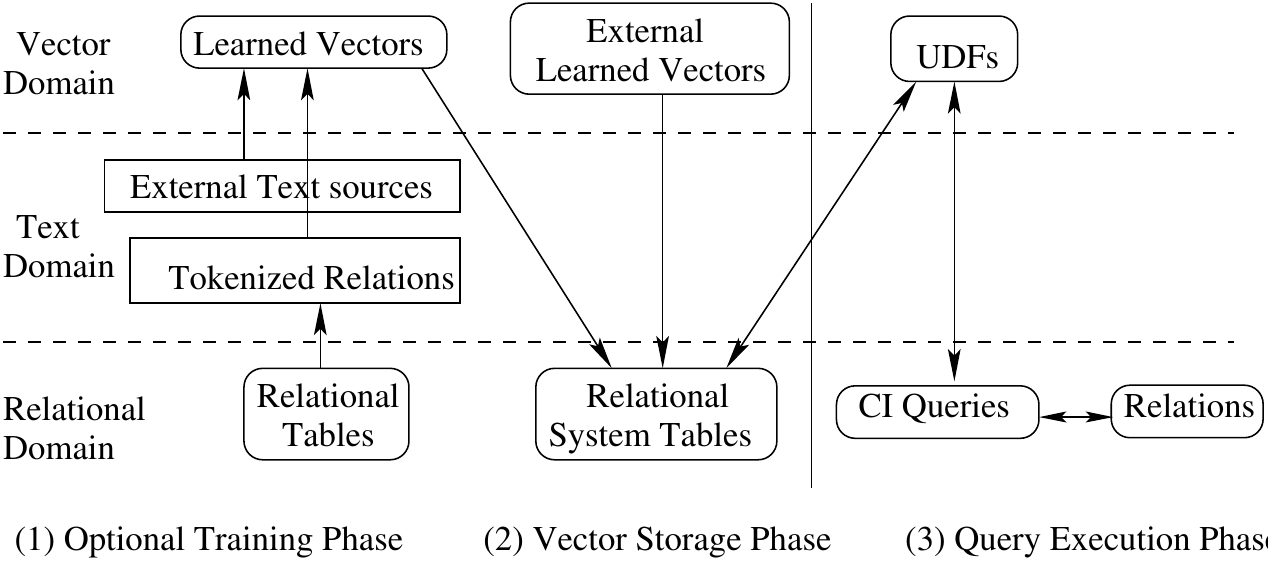}
                \caption{Phases in executing a cognitive intelligence query}
        \end{center}
\label{fig:exec}
\end{figure}

Figure 12 presents the three key phases in the execution
flow of a CI database. The first, optional, training phase takes place when
the database is used to train the model. Using the approaches
described in Section~\ref{sec:vectors}, the database table data is
first tokenized into a meaningful text format and then used to
generate a set of vectors (phase 1). Alternatively, this phase can also use
an external text repository (e.g., Wikipedia) to train the vectors.  Following vector training, 
the resultant vectors are stored in a relational
system  table (phase 2). At runtime, the SQL query execution engine
uses various UDFs that fetch the trained vectors from the system table as needed 
and answer  CI queries (phase 3).

The training phase is optional, and can be
executed in background as a batch process. If vectors are learned
using the database data, as the database evolves, the vectors need be
updated. This can be done by training anew every so often. Alternatively, or in conjunction, the
vectors may be maintained over time. The critical issue is the
introduction of new tokens (that do not have associated vectors). One
way of handling this is to first associate these with an \emph{average
vector} whose entries are very small (positive and negative)
values. Then, initiate training for a short duration (in the case of
word2vec) where the \emph{older} vectors are not allowed to change (or, their
changes are multiplied by a very small positive fraction) and changes to
the new vectors are amplified. The net effect is that the new vectors
are assigned reasonable vector values. A detailed discussion of this
and similar techniques is deferred to a subsequent publication.

\subsection{CI Queries over a DBLP-based Database}

We have implemented an initial prototype of a CI database using the
Apache Spark infrastructure. The prototype uses UDFs programmed in Python.
The main UDFs are \\
\texttt{cosineDistance()}, \texttt{proximityMax()} and
\texttt{proximityAvg()}. All three UDFs rely on functions for
computing the dot product and length of vectors. For our evaluation,
we used the publicly available DBLP bibliography dataset. From the
source XML file, we extracted bibliographic information on some of the
papers that appeared in SIGMOD and VLDB conferences and generated a CSV file in which every row
corresponds to a separate entry for every author (i.e., a paper with
multiple authors gets split into multiple, almost duplicate, rows). The CSV file has the
following fields: PaperID (integer), Author (string), Title (char
array), and Conference (string). To ease the training process, we
converted author names to single words (e.g., Jim Gray to
\texttt{Jim\_Gray}), and also appended the year to a conference name
(e.g., SIGMOD 2002 to \texttt{SIGMOD\_2002}). This CSV file was then
used to populate a relational table in our Spark environment.

For training the vectors, the CSV file was first processed to remove the
commas and a text file containing tokens was created. This text file
was then used in a standard word2Vec application to generate 200 
dimension vectors. The vectors were then loaded into an in-memory
Python structure, containing the relevant words (tokens), and 
with each word its associated vector as an array. Once the vectors were
loaded, the system was ready to perform SQL queries (we omit the
details of starting a Spark SQL environment). For our experiments, we
used a subset of the CSV data to populate the relational tables. Due to space
limitations, in this paper we focus only on the similarity queries.

\begin{table}[htbp]
\centering
{\small
\begin{tabular}{||l|l|l|l||}
\hline
Token & Cosine Distance \\ \hline
Multiversion & 0.437362 \\ \hline
Abraham\_Silberschatz & 0.430781 \\\hline
Timestamp-Based & 0.418086 \\ \hline
Henry\_F.\_Korth & 0.415660 \\ \hline
Non-Two-Phase & 0.412105\\ \hline
Admission & 0.376440 \\ \hline
Ambients & 0.374265 \\ \hline
Laurent\_Amanton & 0.370771 \\ \hline
Daniel\_R.\_Ries & 0.364749 \\ \hline
Transaction & 0.361606 \\ \hline
\end{tabular}
}
\caption{Tokens related to \texttt{Concurrency} sorted by their cosine distances}
\label{tab:related-tokens}
\end{table}

Before discussing  CI queries, let us first get a taste of the vector-implied
relationships among the tokens. Table~\ref{tab:related-tokens} presents of a list of the top 10 tokens that
the word2Vec produced vectors classify as being semantically closest to the input
token \texttt{Concurrency}. The semantic similarity is measured using the
cosine distance between the 200-dimension vectors of the input token,
and the tokens from the input text
data. The tokens listed in Table~\ref{tab:related-tokens} are close
as they co-occur with the input token (\texttt{Concurrency)} in the token
sequence used for vector training (e.g., in a paper title authored by that
person). Thus, these tokens contribute to the overall meaning of the
token \texttt{Concurrency}.

\begin{figure}[htbp]
\hrule
{\small \tt{
\ix \\
 SELECT X.Author, X.Title,  X.Conference \\
cosineDistance(X.Author,``XML'') AS cosineDistance FROM papers X \\
 WHERE  cosineDistance(X.Author, ``XML'') > 0.3\\
 ORDER BY cosineDistance DESC\\
  LIMIT 5\\
}
\hrule
}
\caption{Example of finding authors based on an input topic
  (\texttt{XML})}
\label{fig:topic}
\end{figure}

Next, we demonstrate the use of vector-based similarities for answering
novel CI queries. The first query (Figure~\ref{fig:topic}) aims to
find authors based on an input topic (e.g., \texttt{XML}) and return
the corresponding papers and conferences. The query uses the \texttt{cosineDistance()} UDF
to compare the vectors of all author tokens from the database against
the vector of the \texttt{XML} token. Whereas we demonstrate the query
with an explicit constant (\texttt{XML}), clearly it can be passed to
the query through a program variable. The top 5 nearest authors are
then selected and their papers, with corresponding conferences, are
returned. Table~\ref{tab:topic}(a) presents the results of the query. It
is important to note that the token \texttt{XML} is compared
against the author tokens. As none of the authors had \texttt{XML} in
their names, the matching is purely semantic, based on the cosine
distance between corresponding vectors.

\begin{table*}
\centering
{\small
\begin{tabular}{||l|c||}
\hline
\multicolumn{1}{||c|}{\textbf{(a) Query Results for finding authors based on an input topic}} & Cos. Dist.\\ \hline
Yannis\_Papakonstantinou, Storing and querying XML data using
denormalized relational databases. VLDB\_J.\_2005 & 0.383547 \\ \hline
 Jayavel\_Shanmugasundaram, Efficiently publishing relational data
as XML documents., VLDB\_J.\_2001 & 0.383366 \\ \hline
Vassilis\_J.\_Tsotras, Supporting complex queries on multiversion
XML documents., ACM\_Trans.\_Internet\_Techn.\_2006 & 0.380822 \\ \hline
Shu-Yao\_Chien, Efficient Structural Joins on Indexed XML
Documents., VLDB\_2002 & 0.367735 \\ \hline
Vikas\_Arora, Effective and efficient update of xml in
RDBMS., SIGMOD\_Conference\_2007 & 0.364807 \\ \hline \hline
\multicolumn{1}{||c|}{\textbf{(b) Query Results for finding related authors given an input author}} & Cos. Dist.\\ \hline
Vivek\_R.\_Narasayya,Variance aware optimization of parameterized
queries.,SIGMOD\_Conference\_2010 & 0.671002 \\ \hline
Venkatesh\_Ganti,Ranking objects based on
relationships.,SIGMOD\_Conference\_2006 & 0.495472 \\ \hline
Raghav\_Kaushik,When Can We Trust Progress Estimators for SQL
Queries?,SIGMOD\_Conference\_2005 & 0.483470 \\ \hline
Nicolas\_Bruno,Interactive plan hints for query
optimization.,SIGMOD\_Conference\_2009 & 0.480848 \\ \hline
Sanjay\_Agrawal,Automatic physical design tuning workload as a
sequence.,SIGMOD\_Conference\_2006 & 0.460085 \\ \hline
\end{tabular}
}
\caption{Results from the CI queries to find authors from input topic
  or input author}
\label{tab:topic}
\end{table*}

The second query (Figure~\ref{fig:related-author}) performs a similar task: it finds out authors that
are semantically close to an input author (e.g.,
\texttt{Surajit\_Chaudhuri}). In this case, the vector of the token
\texttt{Surajit\_Chaudhuri} is compared against the vectors of the
authors using the \texttt{cosineDistance()} UDF and the top 5 closest
authors are returned, along with their titles and
conferences. Table~\ref{tab:topic}(b) presents the results of
the query.

\begin{figure}[htbp]
\hrule
{\small \tt{
\ix \\
 SELECT  X.Author, X.Title,  X.Conference \\
cosineDistance(X.Author,``Surajit\_Chaudhuri'') AS cosineDistance FROM papers X \\
 WHERE  cosineDistance(X.Author,``Surajit\_Chaudhuri'') > 0.3\\
 ORDER BY cosineDistance DESC \\
 LIMIT 5\\
}
\hrule
}
\caption{Example of finding related authors given an input author (\texttt{Surajit\_Chaudhuri})}
\label{fig:related-author}
\end{figure}

The final query (Figure~\ref{fig:related-title})  finds papers
with related titles using the \texttt{proximityMax()},
\texttt{proximityAvg() or \newline
proximityTop2Avg()}. In this query, the title of the paper with number
\texttt{471} is ``\texttt{Native Xquery processing 
  in oracle XMLDB}''. In this query, vectors for different tokens in a
title are compared in three ways to identify similar titles:
\texttt{proximityAvg()} selects the title whose average vectors are
close, whereas \texttt{proximityMax()} uses maximum closeness between
individual tokens to select related titles. Finally,
\texttt{proximityTop2Avg()} uses the average of the top two cosine
distances of its two token sequences arguments, thereby 'mixing'
maximum and average. 
As Table~\ref{tab:title}  illustrates these
three UDFs give different sets of answers: \texttt{proximityMax()}
chooses two papers with \texttt{Xquery} in their titles as this is the closest
match (1.0),  \texttt{proximityAvg()} returns a broader set of papers
(note that there is no \texttt{XML} or \texttt{relational} token in
the query title.) \texttt{proximityTop2Avg()} uses two top tokens to
  determine similarity, thus produces a different ordering. For
  example, the the closest title chosen by \texttt{proximityTop2Avg()}
  containts both \texttt{XQuery} and \texttt{XML} tokens.

\begin{figure}[htbp]
\hrule
{\small \tt{
\ix \\
 SELECT X.PaperID, X.Author, X.Title, proximityAvg(X.title, Y.title)
 AS proximityAvg \\
FROM papers X, papers Y WHERE Y.number='471'\\
 AND proximityAvg(X.Title, Y.Title) > 0.3  \\ 
 LIMIT 5\\
}
\hrule
}
\caption{Example of finding related papers using proximityAvg(); an almost identical query uses proximityMax() or proximityTop2Avg()}
\label{fig:related-title}
\end{figure}

\begin{table*}
\centering
{\small
\begin{tabular}{||l|c||}
\hline
\multicolumn{1}{||c|}{\textbf{Query Results using  ProximityAvg()}} & Cos. Dist.\\ \hline
Istvan\_Cseri,Indexing XML Data Stored in a Relational
Database.,VLDB\_2004 & 0.4048 \\ \hline
Rajeev\_Rastogi,DataBlitz A High Performance Main-Memory Storage
Manager.,VLDB\_1998, & 0.3581 \\ \hline
Patricia\_G.\_Selinger,Information Integration and XML in IBM's
DB2.,VLDB\_2002 & 0.3403 \\ \hline
Shinichi\_Morishita,Relational-style XML query.,SIGMOD\_Conference\_2008,
& 0.3316 \\ \hline
Roy\_Goldman,DataGuides Enabling Query Formulation and Optimization in
Semistructured Databases.,VLDB\_1997 & 0.3193 \\ \hline \hline
\multicolumn{1}{||c|}{\textbf{Query Results using ProximityMax()}} & Cos. Dist.\\ \hline
Jerome\_Simeon,Implementing Xquery 1.0 The Galax
Experience.,VLDB\_2003 & 1.0 \\ \hline
Hong\_Su,Semantic Query Optimization in an Automata-Algebra
Combined XQuery Engine over XML Streams.,VLDB\_2004 & 1.0 \\ \hline
Roy\_Goldman,DataGuides Enabling Query Formulation and
Optimization in Semistructured Databases.,VLDB\_1997 & 0.3677 \\ \hline
Bongki\_Moon,FiST Scalable XML Document Filtering by Sequencing
Twig Patterns.,VLDB\_2005 & 0.3661 \\ \hline
Istvan\_Cseri,Indexing XML Data Stored in a Relational
Database.,VLDB\_2004 & 0.35 \\ \hline \hline
\multicolumn{1}{||c|}{\textbf{Query Results using ProximityTop2Avg()}} &
Cos. Dist.\\ \hline
Hong\_Su,Semantic Query Optimization in an Automata-Algebra
Combined XQuery Engine over XML Streams.,VLDB\_2004 & 0.6757\\ \hline
Jerome\_Simeon,Implementing Xquery 1.0 The Galax
Experience.,VLDB\_2003 & 0.6530 \\ \hline
Roy\_Goldman,DataGuides Enabling Query Formulation and
Optimization in Semistructured Databases.,VLDB\_1997 & 0.3532 \\ \hline
Quanzhong\_Li,Indexing and Querying XML Data for Regular Path
Expressions.,VLDB\_2001 & 0.34 \\ \hline
Albrecht\_Schmidt\_0002,XMark A Benchmark for XML Data Management.,
VLDB\_2002 & 0.34 \\ \hline
\end{tabular}
}
\caption{Results from a CI query to find papers with related titles
  using proximityAvg(), proximityMax(), and  proximityTop2Avg()}
\label{tab:title}
\end{table*}

The choice of the proximity function (e.g., \texttt{proximityAvg()},
\texttt{proximityMax()}, or \texttt{proximityTop2Avg()}) as well as
the lower bound on the cosine-distance (e.g., 0.3 or 0.2) is dependent
on the application domain and needs to be fine-tuned based on the
workload characteristics.

\subsection{Performance Issues}

In an end-to-end CI system, there are three spots where performance
can become a concern. First, in the training phase, if the vectors have 
to be trained from the raw database data, depending on the size of the
text used, the vector training time can be very large. However, this
training phase is not necessary as the system can use pre-trained
vectors, and if the vectors need to be trained, the training (or
retraining for new data), can be implemented as a batch
process. Performance of training can be further improved by using
GPUs. Second, the cost of acceessing the trained vectors from the
system tables has to be as minimal as possible as any delay would
impact the performance of runtime query execution (Figure 12).  Access
to the system tables can be improved by building a traditional B$^{+}$-tree
index, with the tokens as keys. Finally, the execution cost of a CI
query is dependent on the performance of the distance
function. In many cases, we may need to compute distances among a
large number of vectors (e.g., for  analogy queries). The distance
calculations can be accelerated either using CPU's SIMD capabilities
or using accelerators such as GPUs. This is a focus of our current
work and we will provide details in subsequent publications.

%% file: concl.tex
\section{Conclusions and Future Work}
\label{sec:concl}

We describe the enhanced querying power resulting from having each
database text entity associated with a vector that captures its
syntactic as well as its semantic characteristics.  Vectors are
obtained by a learning device operating on the database-derived text.  
We use these vectors to enhance database querying capabilities. In
essence, these vectors provide another way to look at the database,
almost orthogonal to the structured relational regime, as
vectors enable a dual view of the data: relational and meaningful
text. We thereby introduce and explore a new class of queries called
\emph{cognitive intelligence (CI)} queries that extract information
from the database based, in part, on the relationships encoded by
these vectors.

We outline a tokenization process and extensions thereof. These
extensions increase the learning precision. We show how SQL
user defined functions (UDFs) can take advantage of these vectors in
querying. In particular, we define some very basic UDFs such as
\texttt{proximityMAX()}, \texttt{proximityAVG()}, and \texttt{proximityTop2Avg()}. SQL queries using
these UDFs implement CI queries. However, we can make CI queries more
expressive by extending standard SQL. These features enable \emph{query navigation} with little
knowledge of the database schema. In large installations, with hundreds
of relations, this is a very useful extension.

We  implemented a prototype system (on top of Apache Spark \cite{Spark}) to
exhibit the power of CI queries. This power goes far beyond text
extensions to relational systems due to the information encoded in
vectors. We also considered various important aspects beyond the basic scheme, including
using a collection of views derived from the database to focus on a
domain of interest, utilizing vectors and text from external sources,
and maintaining vectors as the database evolves. 

We are currently extending our infrastructure to support more complex query patterns. We are also working on accelerating vector training, vector distance
calculations using GPUs, and developing new techniques for incremental
vector training. Although we used an academic scenario (DBLP) to demonstrate 
our ideas, we believe CI queries are applicable to a broad class of
application domains including healthcare, bio-informatics, document
searching, retail analysis, and data integration. We are currently
working on applying the CI capabilities to some of these domains. One
important direction of research is establishing agreed-upon benchmarks
for evaluating and ranking CI systems.  Such benchmarks exist in NLP
and in some database areas such as transaction processing. This is a
non-trivial task as such benchmarks need be of a significant size on
the one hand and may  require  manual construction on the other
hand. One possible direction here is to use crowdsourcing for building
the benchmarks.

%% file: appendix.tex
\section{Appendix}
\label{sec:appendix}

word2vec is a software tool for associating vocabulary words with vectors. The tool  comes in two flavors, continuous bag-of-words (CBOW) and continuous skip-gram (SG).  Intuitively, in CBOW, each word $w$ is represented by two vectors
$V_w$ and $V^{\prime}_w$. The algorithm scans a window of text of
 a pre-determined size $d$, and attempts, by computing a
gradient, to \emph{move} the $V_w$ vectors of the words  $w$ in the window
(except the center word)
\emph{towards} the $V^{\prime}_c$ vector of the window's center word $c$.
SG has a similar but a bit more costly mechanism.
word2vec can employ a technique called {\em negative sampling} (NS) in which examples
having (with high probability, by picking randomly) \emph{no connection} to the
text window  words are presented to the learning device as \emph{negative examples}.
Intuitively, the window words vectors are to move \emph{away} from
the primed vector of a negative example.
The inner workings of word2vec are  explained in great detail in Xin Rong's paper \cite{DBLP:Rong14-v2}.
It turns out that word vectors carry semantic information about various \emph{roles} a word plays. 
For example, a king has roles as a human, male, ruler, monarch,
dictator and others.
There are a few explanations
\cite{goldberg:corr-abs-1402-3722, arora:corr-abs-1502-03520} for this phenomenon od role capturing.

There are other vector construction methods, most notably GloVe~\cite{pennington:glove14} and the method of Arora et. al~\cite{pennington:glove14, arora:corr-abs-1502-03520}.
Vector construction methods have a few characteristics in common:
\begin{itemize}
 \item	The context is usually Natural Language Processing (NLP).
One exception is \cite{perozzi:kdd14} that uses word2vec to produce \emph{synthetic features}
for graph nodes to be later on used by classifiers.
\item
In scanning methods, learning takes place by presenting examples from a
very large corpus of Natural Language (NL) sentences.
The examples may be presented to a  neural network or a learning device such as word2vec.
Examples are usually based on a sliding window through the text.


\item  In matrix methods (e.g., \cite{levy:nips14} and GloVe), text characteristics are extracted into a
  matrix form and an optimization method is utilized to minimize a
  function expressing the word-oriented desired word vector
  representation.

%

 \item	One can manipulate these vectors in order to answer  analogy questions,
cluster similar words based on a similarity measure such as cosine distance by
using a clustering algorithm,
or use these vectors in other analytical models such as a
classification/regression model for making various predictions.
\end{itemize}